\documentclass[a4paper,twoside]{article}
\newcommand{\affil}[1]{$^{\rm #1}$}
\usepackage[authoryear]{natbib}

\bibpunct{(}{)}{;}{a}{}{,}
\usepackage{graphicx}
\usepackage{amsmath}

\date{} 
\title{\large\bf\flushleft Spectral Classification Using Restricted Boltzmann Machine}
\author{\parbox{\textwidth}{\flushleft
\vspace{-0.5cm}
{\it Chen Fuqiang\affil{A}, Wu Yan\affil{A,C}, Bu Yude\affil{B}, Zhao Guodong\affil{A}}\\
\vspace{0.4cm}
{\small \affil{A}\,College of Electronics \& Information Engineering, Tongji University, Shanghai, 201804, China}\\
{\small \affil{B}\,School of Mathematics and Statistics, Shandong University, Weihai, 264209, China}\\
{\small \affil{C}\,Email: yanwu@tongji.edu.cn}}}
\begin{document}
\twocolumn[
\begin{changemargin}{.8cm}{.5cm}
\begin{minipage}{.9\textwidth}
\vspace{-1cm}
\maketitle
%
%
\small{\bf Abstract:}
In this study, a novel machine learning algorithm, restricted Boltzmann machine (RBM), is introduced. The algorithm is applied for the spectral classification in astronomy. RBM is a bipartite generative graphical model with two separate layers (one visible layer and one hidden layer), which can extract higher level features to represent the original data. Despite generative, RBM can be used for classification when modified with a free energy and a soft-max function. Before spectral classification, the original data is binarized according to some rule. Then we resort to the binary RBM to classify cataclysmic variables (CVs) and non-CVs (one half of all the given data for training and the other half for testing). The experiment result shows state-of-the-art accuracy of 100\%, which indicates the efficiency of the binary RBM algorithm.

\medskip{\bf Keywords:} astronomical instrumentation, methods and techniques---methods: analytical---methods: data analysis---
methods: statistical

\medskip
\medskip
\end{minipage}
\end{changemargin}
]
\small

\section{Introduction}

With the rapid development of both the astronomical instruments and various machine learning algorithms, we can apply the spectral characteristics of stars to classify the stars. A great quantity of astronomical observatories have been built to get the spectra, such as the Large Sky Area Multi-Object Fibre Spectroscopic Telescope (LAMOST) in China. A variety of machine learning methods, e.g., principal component analysis (PCA), locally linear embedding (LLE), artificial neural network (ANN) and decision tree etc., have been applied to classify these spectra in an automatic and efficient way. In this study, we apply a novel machine learning method, restricted Boltzmann machine, to classify the CVs and non-CVs.

CVs are composed of the close binaries that contain a white dwarf accreting material from its companion (Warner 2003). Generally, they are small with an orbital period of 1 to 10 hours. The white dwarf is often called "primary" star, while the normal star is called the "companion" or the "secondary" star. The companion star, which is "normal" like our Sun, usually loses its material onto the white dwarf via accretion.

The three main types of CVs are novae, dwarf novae and magnetic CVs. Magnetic CVs (mCVs) are binary star systems with low mass and also with a Roche lobe-filling red dwarf which "gives" material to a magnetic white dwarf. Polars (AM Herculis systems) and Intermediate Polars (IPs) are two major subclasses of mCVs (Wu 2000). More than a dozen of objects have been classified as AM Her systems. Most of the objects were found to be X-ray sources\footnote{http://ttt.astro.su.se/$\sim$stefan/amher0.html} before classified as AM Her's resorting to optical observations.

Besides, Muno et al. presented a catalog of 9017 X-ray sources identified in \textit{Chandra} observations of a $2\times0.8^\circ$ field around the Galactic center (Muno et al. 2009). And they found that the detectable stellar population of external galaxies in X-rays was dominated by accreting black holes and neutron stars, while most of their X-ray sources may be CVs.
\subsection{Previous work in spectral classification in astronomy}
\indent In 1998, Singh et al. applied principal component analysis (PCA) and artificial neural network (ANN) to stellar spectral classification (Singh, Gulati, \& Gupta 1998) on O to M type stars, where O type stars are the hottest and the letter sequence (O to M) indicates successively cooler stars up to the coolest M type stars. They adopted PCA for dimension reduction firstly, in which they reduced the dimension to 20, with the cumulative percentages larger than 99.9 \%. Then they used multi-layer back propagation (BP) neural network for classification.

In 2006, Sarty and Wu applied two well known multivariate analysis methods, i.e., PCA and discriminant function analysis, to analyze the spectroscopic emission data collected by Williams (1983). By using the PCA method, they found that the source of variation had correlation to the binary orbital period. With the discriminant function analysis, they found that the source of variation was connected with the equivalent width of the $H\beta$ line (Sarty \& Wu 2006).

In 2010, Rosalie et al. applied PCA to analyze the stellar spectra obtained from SDSS (Sloan Digital Sky Survey) DR6 (McGurk, Kimball, \& Ivezi$\acute{\textmd{c}}$ 2010). They found that the first 4 principal components (PCs) could remain enough information of the original data without overpassing the measurement noise. Their work made classifying novel spectra, finding out unusual spectra and training a variety of spectral classification methods etc. not as hard as before.

In 2012, Bazarghan applied self-organizing map (SOM, a kind of unsupervised artificial Neural Network (ANN) algorithm) to stellar spectra obtained from the Jacoby, Hunter and Christian (JHC) library, and the author obtained the accuracy of about 92.4\% (Bazarghan 2012). In the same year, Navarro et al. used the ANN method to classify the stellar spectra with low signal-to-noise ratio (S/N) on the samples of field stars which were along the line of sight toward NGC 6781 and NGC 7027 etc. (Navarro, Corradi, \& Mampaso 2012). They not only trained, but also tested the ANNs with various S/N levels. They found that the ANNs were insensitive to noise and the ANN's error rate was smaller when there were two hidden layers in the architecture of the ANN in which there were more than 20 hidden units in each hidden layer.

In the above, some applications of PCA for dimension reduction and ANN for spectral classification were reviewed in astronomy. Furthermore, SVM and decision trees have also been used for spectral classification in astronomy.

In 2004, Zhang and Zhao applied single-layer perceptron (SLP) and support vector machines (SVMs) etc. for the binary classification problem, i.e., the classification of AGNs (active galactic nucleus) and S \& G (stars and normal galaxies) (Zhang \& Zhao 2004), in which they first selected features using the histogram method. They found that SVM's performance was as good as or even better than that of the neural network method when there were more features chosen for classification. In 2006, Ball et al. applied decision trees to SDSS DR3 (Ball et al. 2006). They investigated the classification of 143 million photometric objects and they trained the classifier with 477,068 objects. There were three classes, i.e., galaxy, star and neither of the former two classes, in their experiment.

From the perspective of feature extraction methods, some researches in spectral classification based on linear dimension reduction technique, e.g., PCA, have been reviewed. Except from linear dimension reduction method, nonlinear dimension reduction technique has also been applied in spectral classification for feature extraction.

In 2011, Daniel et al. applied locally linear embedding (LLE, a well known nonlinear dimension reduction technique) to classify the stellar spectra coming from the SDSS DR7 (Daniel et al. 2011). There were 85,564 objects in their experiment. They found that most of the stellar spectra was approximately a 1d sequence lying in a 3d space. Based on the LLE method, they proposed a novel hierarchical classification method being free of the feature extraction process.
\subsection{Previous application of RBM}
In this subsection, we present some representative applications of the RBM algorithm so far.

In 2007, Salakhutdinov et al. (Salakhutdinov, Mnih, \& Hinton 2007) applied RBM for collaborative filtering, which is closely related to recommendation system in machine learning community. In 2008, Gunawardana and Meek (Gunawardana \& Meek 2008) applied RBM for cold start recommendations. In 2009, Taylor and Hinton (Taylor \& Hinton 2009) applied RBM for modeling motion style. In 2010, Dahl et al. (Dahl et al. 2010) applied RBM to phone recognition on the TIMIT dataset. In 2011, Schluter and Osendorfer (Schluter \& Osendorfer 2011) applied RBM to estimate music similarity. In 2012, Tang et al. (Tang, Salakhutdinov, \& Hinton 2012) applied RBM for recognition and de-noising on some public face databases.

\subsection{Our work}
In this study, we applied the binary RBM algorithm to classify spectra of CVs and non-CVs obtained from the SDSS.

Generally, before applying a classifier for classification, the researchers always preprocess the original data, for example, normalization to get better features and thus to get better performance. Thus, firstly, we normalize the spectra with unit norm\footnote{We say the norm of a vector $\textbf{\textit{x}}=(x_1,\ldots,x_n)$ is unit, if $\sum_ix_i^2=1$.}. Then, to apply binary RBM for spectral classification, we binarize the normalized spectra by some rule which we will discuss in the experiment. Finally, we use the binary RBM for classification of the data, one half of all the given data for training and the other half for testing. The experiment result shows that the classification accuracy is 100\%, which is state-of-the-art. And RBM outperforms the prevalent classifier, SVM, with accuracy of 99.14\% (Bu et al. 2013).

The rest of this paper is organized as follows. In section 2, we review the prerequisites for training restricted Boltzmann machine. In section 3, we introduce the binary RBM and the training algorithm for RBM. In section 4, we present the experiment result. Finally, in section 5, we conclude our work in this study and also present the future work.

\section{Prerequisites}
\subsection{Markov Chain}
A Markov chain is a sequence composed of a number of random variables. Each element in the sequence can transit from one state to another one randomly. Indeed, a Markov chain belongs to a stochastic process (Andrieu et al. 2003). In general, the number of possible states for each element or random variable in a Markov chain is finite. And a Markov chain is a random process without memory. It is the current state rather than the states preceding the current state that can influence the next state of a Markov chain. This is the well known Markov Property (Xiong, Jiang, \& Wang 2012).

Mathematically, a Markov chain is a sequence, $X_1$, $X_2$, $X_3$, $\ldots$, with the following property:
\begin{displaymath}
\begin{aligned}
P(X_{n+1}=x_{n+1}|X_1=x_1, X_2=x_2, \ldots, X_n=x_n)\\
={P(X_{n+1}=x_{n+1}|X_n=x_n),}
\end{aligned}
\end{displaymath}
where the $X_i$ ($i=1,2,\ldots$) is a random variable and it usually can take on finite values for a specific problem in the real world. And all the values as a whole can form a denumerable set $S$, which is commonly called the state space of the Markov chain (Yang et~al. 2009).

Generally, all the probabilities of the transition from one state to another one can be represented as a whole by a transition matrix. And the transition matrix has the following three properties:
\begin{itemize}
\item square: both the row number of the matrix and the column number of the matrix equal the total number of the states that the random variable in the Markov chain can take on;
\item the value of a specific element is between 0 and 1: it represents the transition probability from one state to another one;
\item all the elements in each row sum to 1: The sum of the transition probabilities from any specific one state to all the states equals 1.
\end{itemize}
If the initial vector, a row vector, is $\textbf{\textit{X}}_0$, and the transition matrix is $\textbf{\textmd{T}}$, then after $n$ steps of inference, we can get the final vector $\textbf{\textit{X}}_0\cdot \textbf{\textmd{T}}^n$.

Then we introduce the equilibrium of a Markov chain. If there exists an integer $\tilde{N}$, which renders all the elements in the resulting matrix $\textbf{\textmd{T}}^{\tilde{N}}$ nonzero, or rather, greater than $0$, then we say that the transition matrix is a regular transition matrix (Greenwell, Ritchey, \& Lial 2003). If the transition matrix $\textbf{\textmd{T}}$ is a regular transition matrix, and there exists one and only one row vector $\textbf{\textit{V}}$ satisfying the condition that $\textbf{\textit{v}}\cdot \textbf{\textmd{T}}^n$ approximately equals $\textbf{\textit{V}}$, for any probability vector $\textbf{\textit{v}}$ and large enough integer $n$, then we call the vector $\textbf{\textit{V}}$ as the equilibrium vector of the Markov chain.
\subsection{MCMC}
Markov chain Monte Carlo (MCMC) is a sampling algorithm from a specific probability distribution. For the detailed information of MCMC, the readers are referred to Andrieu et al. (2003). The sampling process proceeds in the form of a Markov chain and the goal of MCMC is to get a desired distribution, or rather, the equilibrium distribution via running many inference steps. The larger the number of iterations is, the better the performance of the MCMC is. And MCMC can be applied for unsupervised learning with some hidden variables or maximum likelihood estimation (MLE) learning of some unknown parameters (Andrieu et al. 2003).
\subsection{Gibbs Sampling}
Gibbs sampling method can be used to obtain a sequence of approximate samples from a specific probability distribution, in which sampling directly is usually not easy to implement. For the detailed information of Gibbs sampling, the readers are referred to Gelfand (2000). The sequence obtained via the Gibbs sampling method can be applied to approximate the joint distribution and the marginal distribution with respect to (w.r.t.) one of all the variables etc. In general, Gibbs sampling method is a method for probabilistic inference.

Gibbs sampling method can generate a Markov chain of random samples under the condition that each of the sample is correlated with the nearby sample, or rather, the probability of choosing the next sample equals to 1 in Gibbs sampling (Andrieu et al. 2003).
\section{RBM}
Considering that RBM is a generalized version of Boltzmann Machine (BM), we first review BM in this section. For the detailed information of BM, the readers are referred to Ackley, Hinton, \& Sejnowski (1985).

BM can be regarded as a bipartite graphical generative model composed of two layers in which there are a number of units with both inter-layer and inner-layer connections. One layer is a visible layer $\textbf{\textit{v}}$ with $m$ binary visible units $v_i$, i.e., $v_i=0 \textrm{ or } v_i=1$ ($i=1,2,\ldots,m$). For each unit in the visible layer, the corresponding value is observable. The other layer is a hidden (latent) layer $\textbf{\textit{h}}$ with $n$ binary hidden units $h_j$. As in the visible layer, $h_j=0 \textrm{ or } h_j=1$ ($j=1,2,\ldots,n$). For each unit or neuron in the hidden layer, the corresponding value is hidden, latent or unobservable, and it needs to be inferred.

The units coming from the two layers of a BM are connected with weighted edges completely, with the weights $w_{ij}$ ($v_i\leftrightarrow h_j$) ($i=1,2,\ldots,m,j=1,2,\ldots,n$). For the two layers, the units within each specific layer are also connected with each other, and also with weights.

For a BM, the energy function can be defined as follows:
\begin{equation}
\begin{aligned}
E(\textbf{\textit{v}},\textbf{\textit{h}})=-\sum_{i,j=1}^{m}v_ia_{ij}v_j-\sum_{i,j=1}^{n}h_id_{ij}h_j \\
{-\sum_{i=1}^{m}\sum_{j=1}^{n}v_iw_{ij}h_j-\sum_{i=1}^{m}v_ic_i-\sum_{j=1}^{n}h_jb_j,}\\
\end{aligned}
\end{equation}
where $a_{ij}$ is the weight of the edge connecting visible units $v_i$ and $v_j$, $d_{ij}$ the weight of the edge connecting hidden units $h_i$ and $h_j$, $w_{ij}$ the weight of the edge connecting visible unit $v_i$ and hidden unit $h_j$. For a RBM, the $b_j$ is the bias for the hidden unit $h_j$ in the following activation function (Sigmoid function $f(x)=sigmoid(x)=1/(1+e^{-x})$)
\begin{displaymath} p(h_j=1|\textbf{\textit{v}})=\frac{1}{1+e^{-b_j-\sum_{i=1}^{m}w_{ij}v_i}}. \end{displaymath}
And in a RBM, the $c_{i}$ is the bias for the visible unit $v_i$ in the following formula:
\begin{displaymath} p(v_i=1|\textbf{\textit{h}})=\frac{1}{1+e^{-c_i-\sum_{j=1}^{n}w_{ij}h_j}}. \end{displaymath}
Then for each pair of a visible vector and a hidden vector $(\textbf{\textit{v}},\textbf{\textit{h}})$, the probability of this pair can be defined as follows:
\begin{displaymath} p(\textbf{\textit{v}},\textbf{\textit{h}})=\frac{e^{-E(\textbf{\textit{v}},\textbf{\textit{h}})}}{PF}, \end{displaymath}
where the denominator $PF$ in the fraction (a partition function) is:
\begin{equation} PF=\sum_{\widetilde{\textbf{\textit{v}}},\widetilde{\textbf{\textit{h}}}}p(\widetilde{\textbf{\textit{v}}},\widetilde{\textbf{\textit{h}}}). \end{equation}
Besides, a RBM is a graphical model with the units for both layers not connected within a specific layer, i.e., there are only connections between the two layers for the RBM (Hinton \& Salakhutdinov 2006). Mathematically, for a RBM, $a_{ij}=0$ for $i,j=1,2,\ldots,m$ and $d_{ij}=0$ for $i,j=1,2,\ldots,n$. Thus, the states of all the hidden units $h_j$'s are independent given a specific visible vector $\textbf{\textit{v}}$ and so are the visible units $v_i$'s given a specific hidden vector $\textbf{\textit{h}}$. Then we can obtain the following formula:
\begin{displaymath} p(\textbf{\textit{h}}|\textbf{\textit{v}})=\prod_j p(h_j|\textbf{\textit{v}}) \textrm{  ~~~  and  ~~~ }   p(\textbf{\textit{v}}|\textbf{\textit{h}})=\prod_ip(v_i|\textbf{\textit{h}}). \end{displaymath}
\subsection{Contrastive Divergence}
Contrastive Divergence (CD) is proposed by Hinton and it can be used to train RBM (Hinton, Osindero, \& Teh 2006). Initially, we are given $v_i$ ($i=1,2,\ldots,m$), then we can obtain $h_j$ ($j=1,2,\ldots,n$) by the sigmoid function given in the above. And the value of $h_j$ is determined by comparing a random value $r$ ranging from $0$ to $1$ with the probability $p(h_j=1|\textbf{\textit{v}})$. Then we can reconstruct $\textbf{\textit{v}}$ by $p(v_i=1|\textbf{\textit{h}})$.

We can repeat the above process backward and forward until the reconstruction error is small enough or it has reached the maximum number of iterations which is set beforehand. To update the weights and biases in a RBM, it is necessary to compute the following partial derivative:
\begin{equation} \frac{\partial\log p(\textbf{\textit{v}},\textbf{\textit{h}})}{\partial w_{ij}}=E_{data}[v_ih_j]-E_{recon}[v_ih_j], \end{equation}
\begin{equation} \frac{\partial\log p(\textbf{\textit{v}},\textbf{\textit{h}})}{\partial c_i}=v_i-E_{recon}[v_i], \end{equation}
\begin{equation} \frac{\partial\log p(\textbf{\textit{v}},\textbf{\textit{h}})}{\partial b_j}=E_{data}[h_j]-E_{recon}[h_j], \end{equation}
where $E[\star]$ represents the expectation of $\star$, and the subscript 'data' means that the probability is original-data-driven while the subscript 'recon' means that the probability is reconstructed-data-driven.

Then the weight can be updated according to the following rule:
\begin{displaymath} \Delta w_{ij}=\eta(E_{data}[v_i h_j]-E_{recon}[v_i h_j]), \end{displaymath}
where $\eta$ is a learning rate, which influences the speed of convergence. And the biases can be updated similarly.

In equations (3)-(5), $E_{data}[\star]$'s are easy to compute. To compute or inference the latter term $E_{recon}[\star]$, we can resort to MCMC.

\subsection{Free energy and \textbf{Soft-max}}
To apply RBM for classification, we can resort to the following technique. We can train a RBM for each specific class. And for classification, we need the free energy and the soft-max function for help. For a specific visible input vector $\textbf{\textit{v}}$, its free energy equals to the energy that a single configuration must own and it equals the sum of the probabilities of all the configurations containing $\textbf{\textit{v}}$. In this study, the free energy (Hinton 2012) for a specific visible input vector $\textbf{\textit{v}}$ can be computed as follows:
\begin{equation} F(\textbf{\textit{v}})=-[\sum_iv_ic_i+\sum_j\log(1+e^{x_j})],\end{equation}
where $x_j=b_j+\sum_iv_iw_{ij}$.

For a given specific test vector $\textbf{\textit{v}}$, after training the $RBM_c$ on a specific class $c$, the log probability that $RBM_c$ assigns to $\textbf{\textit{v}}$ can be computed according to the following formula:
\begin{displaymath} \log p(\textbf{\textit{v}}|c)=-F_c(\textbf{\textit{v}})-\log PF_c, \end{displaymath}
here the $PF_c$ is the partition function of $RBM_c$. For a specific classification problem, if the total number of classes is small, there will be no difficulty for us to get the unknown log partition function. In this case, given a specific training set, we can just train a "soft-max" model to predict the label for a visible input vector $\textbf{\textit{v}}$ resorting to the free energies of all the class-dependent $RBM_c$'s:
\begin{equation} \log p(label=c|\textbf{\textit{v}})=\frac{e^{-F_c(\textbf{\textit{v}})-\log \widetilde{PF}_c}}{\sum_de^{-F_d(\textbf{\textit{v}})-\log\widetilde{PF}_d}}. \end{equation}
In the above formula Equation (7), all the partition functions $\widetilde{PF}'s$ can be learned by maximum likelihood (ML) training of the "soft-max" function, where the maximum likelihood method is a kind of parameter estimation method generally with the help of the log probability. Here, the "soft-max" function for a specific unit is generally defined in the following form:
\begin{displaymath} p_j=\frac{e^{x_j}}{\sum_{i=1}^{k} e^{x_i}}, \end{displaymath}
and the parameter $k$ means that there are totally $k$ different states that the unit can take on.

For clarity, we show the complete RBM algorithm in the following. The RBM algorithm as a whole based on the CD method can be summarized as follows:
\begin{itemize}
\item Input: a visible input vector $\textbf{\textit{v}}$; the size of the hidden layer $n_h$; the learning rate $\eta$ and the maximum epoch $M_e$;
\item Output: a weight matrix $\textbf{\textmd{W}}$, a biases vector for the hidden layer $\textbf{\textit{b}}$ and a biases vector for the visible layer $\textbf{\textit{c}}$;
\item Training: \\
      Initialization: Set the visible state with $\textbf{\textit{v}}_1=\textbf{\textit{x}}$, and set $\textbf{\textmd{W}}$, $\textbf{\textit{b}}$ and $\textbf{\textit{c}}$ with small (random) values, \\
      For $t=1,\ldots,M_e$, \\
          For $j=1,\ldots,n_h$, \\
              Compute the following value \\
              $p(h_{1j}=1|\textbf{\textit{v}}_1)=sigmoid(b_j+\sum_iv_{1i}W_{ij})$; \\
              Sample $h_{1j}$ from the conditional distribution $P(h_{1j}|\textbf{\textit{v}}_1)$ with Gibbs sampling method; \\
          End \\
          For $i=1,2,\ldots,n_v$, //Here, the $n_v$ is the size of the visible input vector $\textbf{\textit{v}}$ \\
              Compute the following value \\
              $p(v_{2j}=1|\textbf{\textit{h}}_1)=sigmoid(c_i+\sum_jW_{ij}h_{1j})$; \\
              Sample $v_{2i}$ from the conditional distribution\\
              $P(v_{2i}|\textbf{\textit{h}}_1)$ with Gibbs sampling method; \\
          End \\
          For $j=1,\ldots,n_h$, \\
              Compute the following value \\
              $p(h_{2j}=1|\textbf{\textit{v}}_2)=sigmoid(b_j+\sum_iv_{2i}W_{ij})$; \\
          End \\
      Update the parameters: \\
      $\textbf{\textmd{W}}=\textbf{\textmd{W}}+\eta [P({h_1=1}|\textbf{\textit{v}}_1)\textbf{\textit{v}}_1-P({h_2=1}|\textbf{\textit{v}}_2)\textbf{\textit{v}}_2]$; \\
      $\textbf{\textit{c}}=\textbf{\textit{c}}+\eta (\textbf{\textit{v}}_1-\textbf{\textit{v}}_2)$; \\
      $\textbf{\textit{c}}=\textbf{\textit{c}}+\eta [P({h_1=1}|\textbf{\textit{v}}_1)-P({h_2=1}|\textbf{\textit{v}}_2)]$; \\
      End \\
\end{itemize}
For classification, after training the RBM using the above algorithm, we need to compute the free energy function by Equation (6) and then we can assign a label for the sample $\textbf{\textit{v}}$ with Equation (7).
\section{Experiment}
\subsection{Data description}
There have been a large amount of surveys in astronomy. SDSS is one of those surveys and it is one of the most not only ambitious but also influential ones (The official website of SDSS is http://www.sdss.org/). The SDSS has begun collecting data since 2000. From 2000 to 2008, the SDSS collected deep and multi-color images containing no less than a quarter of the sky and it also created 3d maps for over 930,000 galaxies and also for over 120,000 quasars. Data Release 7 (DR7) is the seventh major data release and it provides spectra and redshifts etc. for downloading.

All the data used in our experiment is coming from the SDSS. All the samples in the entire data set are divided into two classes, one class composed of non-CVs while the other class composed of CVs. There are totally $6818$ non-CVs and $208$ CVs in our data set. Each sample is composed of $3522$ variables, or rather, spectral components. Among the total 6818 non-CVs, there are 1,559 belonging to Galaxies, 3,981 belonging to Stars and the remaining 1,278 belonging to QSOs (Quasi-stellar objects)\footnote{For detail, the readers are referred to the official website of SDSS DR7: http://www.sdss.org/dr7/}.

\begin{figure}[h]
\begin{center}
\includegraphics[scale=0.35, angle=0]{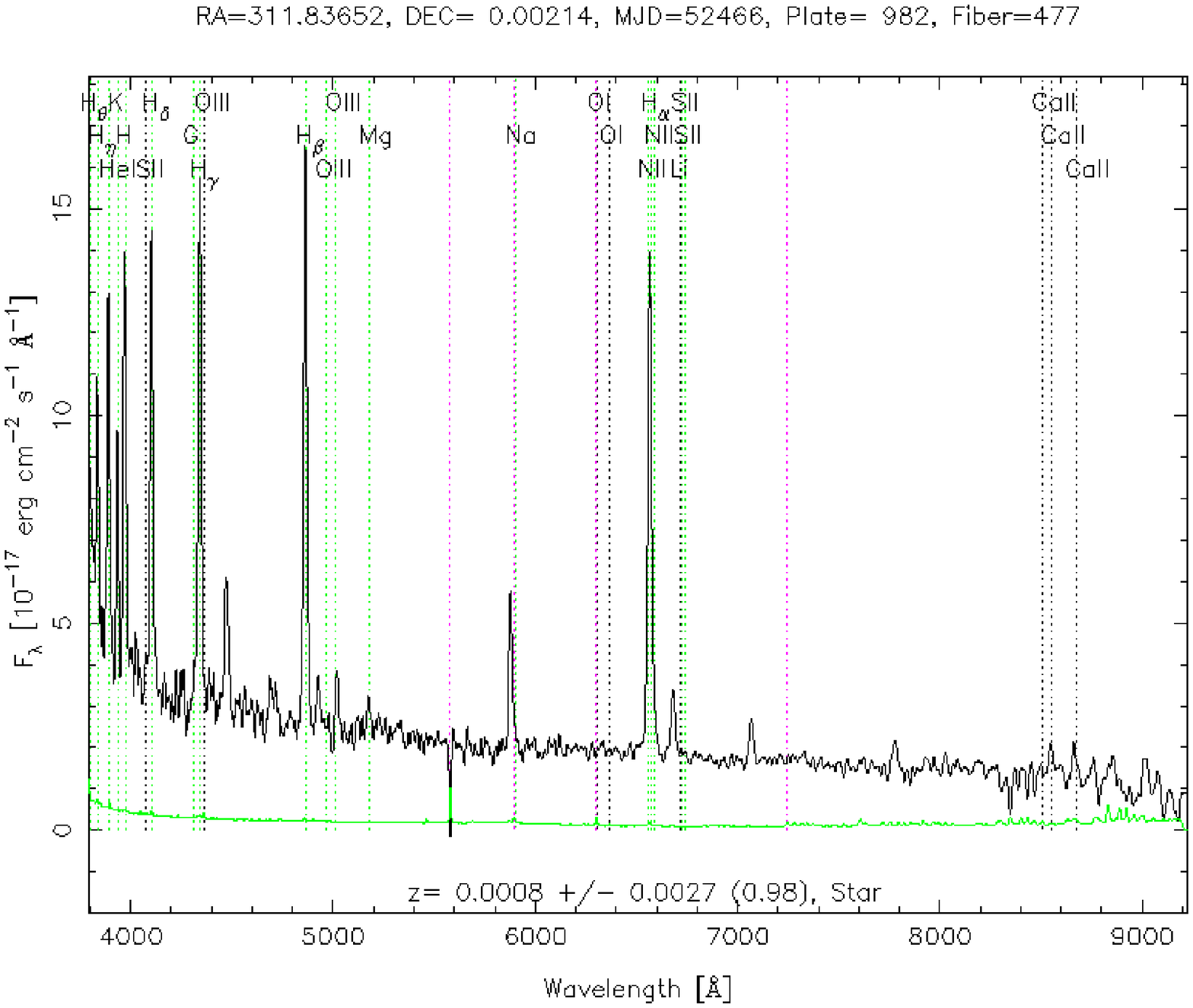}
\caption{ Spectrum of a cataclysmic variable star. The online version is available at: http://cas.sdss.org/dr7/en/tools/explore/
obj.asp?id=587730847423725902}\label{Figure1}
\end{center}
\end{figure}

\begin{figure}[h]
\begin{center}
\includegraphics[scale=0.25, angle=0]{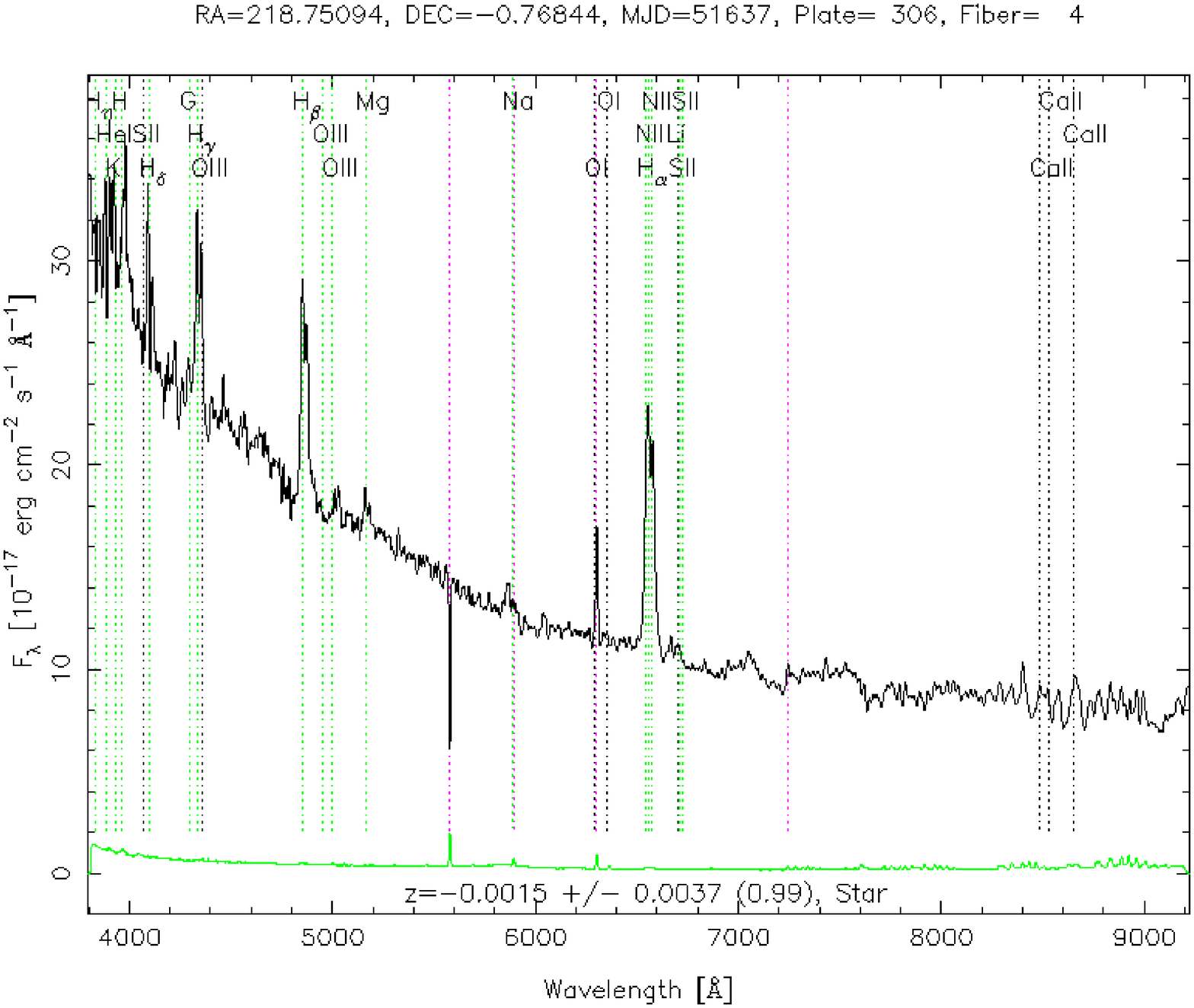}
\caption{ Spectrum of a cataclysmic variable star in our data set.}\label{Figure2}
\end{center}
\end{figure}

In the following, we show the CVs in detail in our experiment. It is common that there will be transparent Balmer absorption lines in their spectra when the CVs outburst. A representative spectrum of the CV from the SDSS is shown in Figure \ref{Figure1}. Much work has been done on the CVs for ages. Without high-tech, the earlier researches are focused on the optical characteristics of the spectrum. Then with the help of the high-tech astronomical instruments, the multi-wavelength studies of the spectrum become to be true and the astronomers can obtain much more information about the CVs than before (Bu et al. 2013). From 2001 to 2006, Szkody et al. had been using the SDSS to search for CVs. The CVs in our data set are from their studies (Szkody et al. 2002, 2003, 2004, 2005, 2006, 2007), and we are deeply grateful to their researches. For clarity, we show the number of the CVs they found using the SDSS in Table \ref{table2}. And the spectrum of a CV in our data set is shown in Figure \ref{Figure2}.
\begin{table}[h]
\caption{The number of the CVs that Szkody et al. searched using the SDSS.}\label{table2}
\begin{center}
\begin{tabular}{ll}
\\ \hline
Paper                      &  \# of CVs\\
\hline
Szkody et al. 2002    &     22  \\
\hline
Szkody et al. 2003    &    42  \\
\hline
Szkody et al. 2004    &    36  \\
\hline
Szkody et al. 2005    &    44  \\
\hline
Szkody et al. 2006    &    41  \\
\hline
Szkody et al. 2007    &    28  \\
\hline
\end{tabular}
\end{center}
\end{table}

In our experiment, we chose randomly half of the whole data for training and the remaining half for testing for both non-CVs and CVs. In detail, for non-CVs, half of the total 6818 samples (i.e. 3414) were randomly chosen to train the RBM classifier and the remaining half to test the RBM classifier. Similarly, for CVs, half of all the 208 samples (i.e. 104) were randomly chosen to train the RBM classifier and the remaining half to test the RBM classifier. To explain it clearly, we showed the data used for training and testing the RBM classifier in the following table (Table \ref{table1}).
\begin{table}[h]
\begin{center}
\caption{The number of the original data for training and for testing respectively, where the number 3522 is the dimension of the original data.} \label{table1}
\begin{tabular}{lcc}
\\
\hline CV/non-CV &    Train/Test      &     \#$\times$Dim     \\
\hline non-CV    &    Train           &     3414$\times$3522  \\
\hline non-CV    &    Test            &     3414$\times$3522  \\
\hline CV        &    Train           &     104$\times$3522   \\
\hline CV        &    Test            &     104$\times$3522   \\
\hline
\end{tabular}
\medskip\\
\end{center}
\end{table}

\subsection{Parameter chosen}
In this subsection, we present the parameters in our experiment. We chose all the parameters referring to Hinton (2012). The learning rate in the process of updating was set to be 0.1. The momentum for smoothness and to prevent over-fitting was chosen to be 0.5. The maximum number of epochs was chosen to be 50. The weight decay factor, penalty, was chosen to be 2$\times$10$^{-4}$. The initial weights were randomly generated from the standard normal distribution, while the biases vectors $\textbf{\textit{b}}$ and $\textbf{\textit{c}}$ were initialized with $\textbf{\textit{0}}$. For clarity, we present them in the following table (Table \ref{table3}).
\begin{table}[h]
\caption{The parameters in our experiment.}\label{table3}
\begin{center}
\begin{tabular}{ll}
\hline
\multicolumn{1}{c}{\bf Parameter}   & \multicolumn{1}{c}{\bf Value}
\\ \hline
learning rate       &  0.1  \\
momentum            &  0.5  \\
maximum epochs      &  50   \\
number of hidden units        &  100  \\
initial biases vector        &  $\textbf{\textit{0}}$    \\
\hline
\end{tabular}
\end{center}
\end{table}
\subsection{Experiment result}
We first normalized the data to make it have unit $l_2$ norm, i.e., for a specific vector $\textbf{\textit{x}}=[x_1,x_2,\ldots,x_n]$, the $l_2$ norm of the vector satisfies $\sum_ix_i^2=1$. Then we could get two matrixes, one was $\textbf{\textmd{A}}=6818\times3522$ and the other was $\textbf{\textmd{B}}=208\times3522$. Then we found out the maximum element and the minimum element for CVs and non-CVs respectively. Finally, to apply binary RBM for classification, we found a parameter to assign the value of the variable in our experiment with $0$ or $1$, or rather, binarization.

Mathematically, if
\begin{displaymath} S(i,j)-\min S(i,j)<\alpha(\max S(i,j)-\min S(i,j)),\end{displaymath}
then we set $S(i,j)$ with $0$, otherwise we set $S(i,j)$ with $1$. Here we used $S(i,j)$ (after binarization) to denote the element of the matrix $\textbf{\textmd{A}}$ and $\textbf{\textmd{B}}$ in the $i^{th}$ row and the $j^{th}$ column. The parameter $\alpha$ satisfied $0<\alpha<1$. To investigate the influence of the parameter $\alpha$ on the final performance of the binary RBM algorithm, we first chose it to be $1/2$ heuristically. Then we chose it to be $1/3$. The experiment result shows that the classification accuracy is 100\%, which is state-of-the-art and it outperforms the prevalent classifier SVM (Bu et al. 2013).

For clarity, we show the result in Table \ref{table4}, in which we also show the performance the binary RBM algorithm based on other values for the variable $\alpha$. From Table \ref{table4}, we can see that the classification accuracy is 97.2\% when $\alpha$ = $1/2$. However, almost all of the CVs for testing is labeled as non-CVs.

Table \ref{table4} shows the classification accuracy computed by the following formula:
\begin{equation} Acc=\frac{\sum[\hat{\textbf{\textit{y}}}==\textbf{\textit{y}}]}{Card(\textbf{\textit{y}})}, \end{equation}
where $\textbf{\textit{y}}$ is a vector denoting the label of all the test samples. In our experiment, there are 3413 (3409 non-CVs + 104 CVs) test samples.  And "Card($\textbf{\textit{y}}$)" represents the number of elements in vector $\textbf{\textit{y}}$. In Equation (8), the denominator $\hat{\textbf{\textit{y}}}$ is the label of all the test samples predicted by Equation (7), in which $c=+1$ or $c=-1$. In this paper, $c=+1$ means that the sample belongs to non-CVs, while $c=-1$ means that the sample belongs to CVs\footnote{You can use any two different integers to represent the labels of the samples belonging to non-CVs and CVs, and this does not impact the result of the experiment.}. And $\sum[\hat{\textbf{\textit{y}}}==\textbf{\textit{y}}]$ means the total number of equal elements in vector $\textbf{\textit{y}}$ and vector $\hat{\textbf{\textit{y}}}$.
\begin{table}[h]
\caption{The classification accuracy with different $\alpha's$.}\label{table4}
\begin{center}
\begin{tabular}{ll}
\hline
\multicolumn{1}{c}{\bf $\alpha$}  &\multicolumn{1}{c}{\bf Accuracy}
\\ \hline
$1/5$         &  97\%  \\
\hline
$2/5$         &  \textbf{100\% }\\
\hline
$3/5$         &  97\%  \\
\hline
$4/5$         &  97\%  \\
\hline
$1/4$         &  97\%   \\
\hline
$1/2$         &  97.2\% \\
\hline
$3/4$         &  97\%  \\
\hline
$1/3$         &  \textbf{100\%}  \\
\hline
$2/3$         &  97\%   \\
\hline
\end{tabular}
\end{center}
\end{table}
\section{Conclusion and future work}
Restricted Boltzmann machine is a bipartite generative graphical model which can extract features representing the original data well. By introducing free energy and soft-max function, RBM can be used for classification. In this paper we apply restricted Boltzmann machine (RBM) for spectral classification of non-CVs and CVs. And the experiment result shows that the classification accuracy is 100 \%, which is the state-of-the-art and outperforms the rather prevalent classifier, SVM.

Since RBM is the building block of deep belief nets (DBNs) and deep Boltzmann machine (DBM), then we can infer that deep Boltzmann machine (Salakhutdinov \& Hinton 2009) and deep belief net can also perform well on spectral classification, which is our future work.
%


%

\section*{Acknowledgments} 
The authors are very grateful to the anonymous reviewer for a thorough reading, many valuable comments and rather helpful suggestions. The authors thank the editor Bryan Gaensler a lot for the helpful suggestions on the organization of the manuscript. The authors also thank Jiang Bin for providing the CV data.



\end{document}